\documentclass[11pt,a4paper]{article}
\usepackage[final]{emnlp2021}
\usepackage{graphicx}
\usepackage{verbatim}
\usepackage[inline]{enumitem}
\usepackage{times}
\usepackage{latexsym}
\usepackage{lipsum}
\usepackage{amsfonts} 
\usepackage{makecell}
\usepackage{booktabs}
\usepackage{mathtools}
\usepackage{import}

\usepackage{microtype}
\usepackage{ulem}
\usepackage{amsmath}

\usepackage{array}
\usepackage{soul}
\usepackage{colortbl}
\usepackage{float}
\usepackage{multirow}
\usepackage{multicol}
\setlist[itemize]{noitemsep, topsep=0pt}

\definecolor{DarkGreen}{rgb}{0.0, 0.4, 0}
\definecolor{DarkYellow}{rgb}{0.4, 0.2, 0.0}
\definecolor{DarkPurple}{rgb}{0.44, 0.16, 0.39}
\definecolor{DarkRed}{rgb}{0.6,0,0}
\definecolor{DarkBlue}{rgb}{0,0,0.6}
\colorlet{LightYellow}{white!80!yellow}
\colorlet{LightRed}{white!80!red}
\colorlet{LightPurple}{white!80!purple}
\colorlet{LightBlue}{white!80!blue}
\colorlet{LightGreen}{white!80!green}

\DeclareRobustCommand{\hlred}[1]{{\sethlcolor{LightRed}\hl{#1}}}
\DeclareRobustCommand{\hlyellow}[1]{{\sethlcolor{LightYellow}\hl{#1}}}

\DeclareRobustCommand{\hlblue}[1]{{\sethlcolor{LightBlue}\hl{#1}}}
\DeclareRobustCommand{\hlgreen}[1]{{\sethlcolor{LightGreen}\hl{#1}}}

\makeatletter
\newcommand{\printfnsymbol}[1]{%
  \textsuperscript{\@fnsymbol{#1}}%
}
\makeatother

\title{\textsc{BiSECT}: Learning to Split and Rephrase Sentences with Bitexts}

\author{Joongwon Kim$^1$\thanks{\hspace{4pt} Equal contribution.}, \ Mounica Maddela$^2$\printfnsymbol{1}, Reno Kriz,$^{1,3}$ Wei Xu,$^2$ Chris Callison-Burch$^1$\\
$^{1}$Department of Computer and Information Science, University of Pennsylvania, \\
$^{2}$School of Interactive Computing, Georgia Institute of Technology \\
$^{3}$ Human Language Technology Center of Excellence, Johns Hopkins University \\
{\small {\tt \{jkim0118, ccb\}@seas.upenn.edu}, {\tt \{mmadela3, wei.xu\}@cc.gatech.edu}, {\tt rkriz1@jh.edu}}\\
}

\date{}

\begin{document}
\maketitle
\begin{abstract}
An important task in NLP applications such as sentence simplification is the ability to take a long, complex sentence and split it into shorter sentences, rephrasing as necessary.  
We introduce a novel dataset and a new model for this `split and rephrase' task. Our \textsc{BiSECT} training data consists of 1 million long English sentences paired with shorter, meaning-equivalent English sentences. We obtain these by extracting 1-2 sentence alignments in bilingual parallel corpora and then using machine translation to convert both sides of the corpus into the same language. \textsc{BiSECT} contains higher quality training examples than previous Split and Rephrase corpora, with sentence splits that require more significant modifications. We categorize examples in our corpus, and use these categories in a novel model that allows us to target specific regions of the input sentence to be split and edited. Moreover, we show that models trained on \textsc{BiSECT} can perform a wider variety of split operations and improve upon previous state-of-the-art approaches in automatic and human evaluations.\footnote{Our code and data are available at \url{https://github.com/mounicam/BiSECT}.}

\end{abstract}

\section{Introduction}


Understanding long and complex sentences is challenging for both humans and NLP models.  NLP tasks like machine translation \cite{pouget2014overcoming,koehn2017six} and dependency parsing \cite{mcdonald2011analyzing} tend to perform poorly on long sentences. Text simplification  \cite{zhu2010monolingual,xu2015problems} is often formulated with a specific step to break longer sentences into shorter sentences.  This task is referred to as  Split and Rephrase \cite{narayan2017split}.  

Several past efforts have created Split and Rephrase training sets, which consist of long, complex input sentences paired with multiple shorter sentences that preserve the meaning of the input sentence.
\newcite{narayan2017split} introduced the \textsc{WebSplit} corpus based on decomposing a long sentence into RDF triples (a form of semantic representation), and generating shorter sentences from subsets of these triples. However, the reliance on RDF triples and a limited vocabulary results in unnatural expressions \cite{botha2018learning} and repeated syntactic patterns \cite{zhang2020small}.

\begin{figure}[bt]
    \centering
    \includegraphics[width=0.9\linewidth]{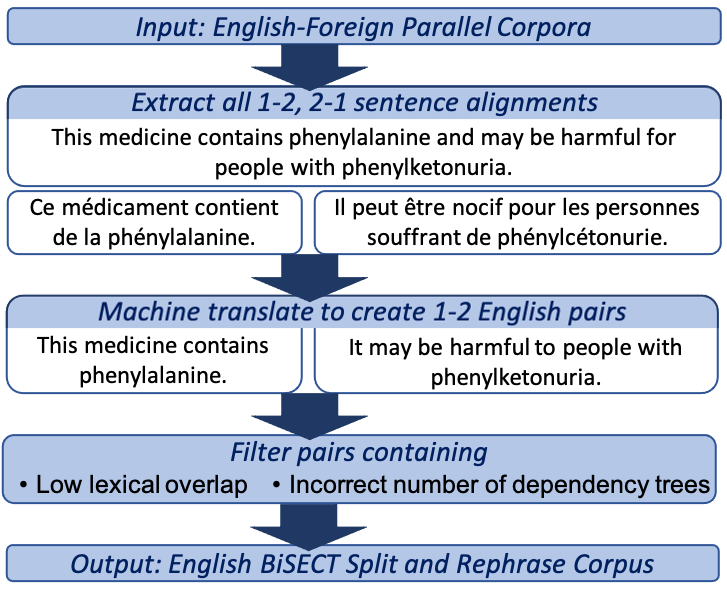}
    \caption{The process of creating the English \textsc{BiSECT} Split and Rephrase corpus.}
    \label{fig:data_collection}
\end{figure}

More recently, the \textsc{WikiSplit} corpus \cite{botha2018learning} was introduced. It contains one million training examples of sentence splitting  that were  mined from the revision history of English Wikipedia. While this yields an impressive number of training examples, the data are often quite noisy, with around 25\% of \textsc{WikiSplit} pairs containing significant errors (detailed in \S \ref{sec:manual_quality}). This is because Wikipedia editors are not only trying to split a sentence, but also often simultaneously modifying the sentence for other purposes, which results in changes of the initial meaning. 

In this paper, we introduce a novel methodology for creating Split and Rephrase corpora via bilingual pivoting \cite{wieting2018paranmt,hu2019large}. Figure \ref{fig:data_collection} demonstrates the process. First, we extract all 1-2 and 2-1 sentence-level alignments \cite{gale1993program} from bilingual parallel corpora, where a single sentence in one language aligns to two sentences in the other language.  We then machine translate the foreign sentences into English.
The result is our \textsc{BiSECT} corpus. 





Split and Rephrase corpora, including \textsc{BiSECT}, contain pairs with variable amounts of rephrasing. Some pairs only edit around the split location, while others require more involved changes to maintain fluency. In this work, we leverage this knowledge by introducing a classification task to predict the amount of rephrasing required, and a novel model that targets that amount of rephrasing.

The main contributions of this paper are:

\begin{itemize}
    \item We introduce \textsc{BiSECT}, the largest multilingual Split and Rephrase corpus.  BiSECT contains 938K English pairs, 494K French pairs, 290K Spanish pairs, and 186K German pairs. 
    \item We show that \textsc{BiSECT} is higher quality than \textsc{WikiSplit}, that it contains a wider variety of splitting operations, and that models trained with our resource produce better output for the Split and Rephrase task.
    \item We introduce a novel classification task to identify the types of sentence splitting outputs based on how much rephrasing is necessary.
    \item We develop a novel Split and Rephrase model that accounts for these classifications to control the amount of rephrasing.
\end{itemize}




\section{Related Work}


The idea of splitting a sentence into multiple shorter sentences was initially considered a sub-task of text simplification \cite{zhu2010monolingual,narayan2014hybrid}. However, the structural paraphrasing required to split a sentence makes for an interesting problem in itself, with many downstream NLP applications. Thus, \newcite{narayan2017split} proposed the Split and Rephrase task, and introduced the \textsc{WebSplit} corpus, created by aligning sentences in WebNLG \cite{gardent2017creating}. 
\textsc{WebSplit} contains duplicate instances and phrasal repetitions \cite{aharoni2018split,botha2018learning}, and most splitting operations can be trivially classified \cite{zhang2020small}, so subsequent Split and Rephrase corpora have been created to improve training \cite{botha2018learning} and evaluation \cite{sulem2018bleu,zhang2020small}. The main work we compare against is \textsc{WikiSplit}, a corpus created by extracting split sentences from Wikipedia edit histories \cite{botha2018learning}. Concurrent work used a subset of \textsc{WikiSplit} to focus on sentence decomposition \cite{gao2021abcd}. While this approach is able to both extract many potential sentence splits and transfer across languages, edited sentences do not 
necessarily have to retain the same meaning. In contrast, our corpus \textsc{BiSECT} is created from aligned parallel documents.

\begin{table*}[ht]
\setlength{\tabcolsep}{4pt}
\centering
\small
\begin{tabular}{l|ccrlrcc}
  \toprule
   \multirow{2}{*}{\textbf{Dataset}} & \multirow{2}{*}{\textbf{Pivot Lang.}}  & \multirow{2}{*}{\textbf{Domain}} & \multicolumn{3}{c}{\textbf{1-2 \& 2-1 Alignments} }  & \multicolumn{2}{c}{\textbf{Length}} \\
   & & & \multicolumn{2}{c}{total (count/\%)} & after filtering  & Long & Split \\
  \midrule
  \textsc{CCAligned}  & fr  & web crawl   & 559,826 & (20.9\%) & 203,780{ }{ }{ }{ }{ } & 36.0 & 20.1\\  
  \textsc{Europarl}  & fr &  European Parliament & 153,220 & (5.72\%) & 57,473{ }{ }{ }{ }{ } & 45.6 & 23.4\\
  \textsc{$10^9$ FR-EN} & fr & newswire & 624,381 & (23.31\%) & 264,203{ }{ }{ }{ }{ } & 41.8 & 22.5\\
  \textsc{ParaCrawl} & fr,de,es,nl,it,pt & web crawl & 1,212,982 & (45.29\%) & 405,612{ }{ }{ }{ }{ } & 38.5 & 19.7\\
  \textsc{UN} & fr,es,ar,ru & United Nations & 113,840 & (4.25\%) & 64,690{ }{ }{ }{ }{ } &  45.5 & 24.4\\ \cmidrule{1-8}
  \textsc{EMEA}  &  fr &  European Medicines Agency & 5,719 & (0.21\%) & 1,056{ }{ }{ }{ }{ } & 34.1 & 19.7\\
  \textsc{JRC-Acquis} & fr,de & European Union  & 8,358 & (0.31\%) & 6,237{ }{ }{ }{ }{ } & 51.8 & 26.6\\
  
\bottomrule
\end{tabular}
\caption{Datasets from OPUS that were used to create the English version of \textsc{BiSECT}. The training set consists of five corpora in the upper part of the table, while the two corpora in the lower part are used for the development and test sets. We also report the token length of the long sentence and that of the individual split sentences.}
\label{tab:OPUS_corpus_breakdown}
\end{table*}

Bilingual corpora is generally leveraged for monolingual tasks with bilingual pivoting \cite{bannard2005paraphrasing}, which assumes that two English phrases that translate to the same foreign phrase have similar meaning. 
This technique was used to create the Paraphrase Database \cite{ganitkevitch2013ppdb,pavlick2015ppdb}, a collection of over 100 million paraphrase pairs, and to improve neural approaches for sentential paraphrasing \cite{mallinson2017paraphrasing,wieting2018paranmt,hu2019parabank,hu2019large} and sentence compression \cite{mallinson2018sentence}. 


In introducing the Split and Rephrase task, \newcite{narayan2017split} also reports the performance of several baseline models, where the strongest is an LSTM-based model. Subsequent works have improved performance using a copy-attention mechanism \cite{aharoni2018split}. We instead start with a BERT-initialized transformer model \cite{rothe2020leveraging}, and train it with an adaptive loss function to emphasize split-based edits. Concurrent work also introduced an additional neural graph-approach for Split and Rephrase \cite{gao2021abcd}.

\section{\textsc{BiSECT} Corpus}

To address the need of Split and Rephrase data that is both meaning preserving and sufficient in size for training, we present the \textsc{BiSECT} corpus.

\subsection{Corpus Creation Procedure}

The construction of the \textsc{BiSECT} corpus relies on leveraging the sentence-level alignments from OPUS \cite{tiedemann2004opus}, a publicly available collection of bilingual parallel corpora over many language pairs.
While most of the translated sentences in OPUS are aligned 1-1, i.e., one sentence in Language $A$ is mapped to one sentence in Language $B$, 
there are many aligned pairs consisting of multiple sentences from either $A$ or $B$. This is a result of natural variation in the process of human translation. Sentence alignment algorithms \cite{gale1993program} match 1-1, 2-1, and 1-2 alignments in bitext. We extract all 1-2 and 2-1 sentence alignments from parallel corpora, where $A$ is English and $B$ is one of several foreign languages. \



Next, the foreign sentences are translated into English using Google Translate's Web API service\footnote{https://pypi.org/project/googletrans/} 
to obtain English sentence alignments between a single long sentence $l$ and two corresponding split sentences $s= (s_1, s_2)$. As the alignment information provided by OPUS is based on the presence of a sentence-breaking punctuation, there are noisy alignments where $l$ contains a pair of sentences instead of one complex sentence. These noisy alignments belong to two categories: two sentences pasted contiguously without any space around the sentence-breaking delimiter and two independent sentences joined by a space without any punctuation. For the first case, we remove $l$ and its corresponding splits when it contains a token with a punctuation after the first two and before the last two alphabetic characters. For the second case, we generate a dependency tree\footnote{We generate dependency trees using Spacy.} for $l$ and discard $l$ if it contains more than one unconnected component.

 
 
 Moreover, we remove the misalignment errors based on lexical and semantic overlap. We compute lexical overlap ratio $r$ as follows:
 
\vspace{-3mm}
\begin{equation*}
\begin{aligned}
r ={} & min\Bigg(\frac{|\mathcal{L}_l\cap \mathcal{L}_{s_1}|}{|\mathcal{L}_{s_1}|}, \frac{|\mathcal{L}_l\cap \mathcal{L}_{s_2}|}{|\mathcal{L}_{s_2}|}, \\
& \hspace{1.01cm} \frac{|\mathcal{L}_l\cap (\mathcal{L}_{s_1}\cup\mathcal{L}_{s_2})|}{|\mathcal{L}_{s_1}\cup\mathcal{L}_{s_2}|} \Bigg),
\end{aligned}
\end{equation*}

\noindent where $\mathcal{L}_l$, $\mathcal{L}_{s_1}$ and $\mathcal{L}_{s_2}$ denote the sets of lemmatized tokens in $l$ and $(s_1, s_2)$, respectively. We consider an aligned pair valid if $r \geq 0.25$ and $l$, $s_1$ and $s_2$ all contain a verb. We discard invalid pairs. We also remove $(l, s)$ pairs with length-penalized BERTScore $< 0.4$ \cite{zhang2020bertscore, maddela-etal-2021-controllable}.\footnote{We also tried to fix the grammatical errors in the $(l, s)$ pairs using GECToR \cite{omelianchuk-etal-2020-gector}. However, GECToR introduced minimal one word changes that did not help in improving the quality of the data.}

We repeat this  process over all available parallel corpora for each English-Foreign language pair, resulting in 938,102 filtered English-English pairs. An important characteristic of BiSECT to note is that its size can be further increased with the addition of new parallel corpora on OPUS, processed in the method described above.

Table \ref{tab:OPUS_corpus_breakdown} breaks down the OPUS corpora and parallel languages used in creating the English version of \textsc{BiSECT}. For the testing set, a different set of corpora is used from the training set to prevent domain overlap. Moreover, the choice of corpus is based on the number of alignments extracted from each corpus. We choose corpora of relatively smaller sizes for development and testing to avoid a loss of size in the training set. To demonstrate our approach can be extended to other languages,
we also create \textsc{BiSECT} corpora for French, Spanish, and German, using English as the pivot language. Corpus statistics of non-English languages are given in Appendix \ref{app:multilingual}.

\setlength{\tabcolsep}{2.5pt}
\begin{table}[bt]
\centering
\small
\begin{tabular}{l|ccccc}
  \toprule
   \multirow{2}{*}{\textbf{Corpus}} & \multirow{2}{*}{\textbf{\#pairs}} & \multirow{2}{*}{\textbf{\#unique}} & \multirow{2}{*}{\textbf{\%new}} & \multicolumn{2}{c}{\textbf{Length}} \\
   &  &  & & Long & Split \\ \midrule
  
  
  \textsc{HSplit-Wiki}$^\dagger$   & 1436 & 359 & 33.9  & 22.6  & 14.3  \\ 
  \textsc{Contract}$^\dagger$   & 659 & 406 & 10.7  & 39.7 & 22.9 \\
  \textsc{Wiki-BM}$^\dagger$  & 720 & 403 & 8.9 & 29.2 & 16.1 \\  
  \textsc{\textsc{WebSplit}v1.0}  & 1.06M & 17k & 32.1  & 34.3 & 30.2 \\
  \textsc{WikiSplit}  & 999K & 999K & 15.5  & 33.4 & 19.0\\
  \textsc{BiSECT} (this work) & 938K & 938K & 34.6  & 40.1 & 20.6  \\
\bottomrule
\end{tabular}
\caption{Comparison of Split and Rephrase corpora. We compute the number of aligned pairs (\textbf{\#pairs}); number of unique long sentences $l$ (\textbf{\#unique}); the percentage of new words added to $s$ compared to $l$ (\textbf{\%new}), and the average token \textbf{Length} of $l$ and that of the individual split sentences. $^\dagger$ marks crowdsourced corpora.}
\label{tab:corpus_statistics}
\end{table}

\begin{table*}[ht!]
\setlength{\tabcolsep}{4pt}
\small
\centering
\begin{tabular}{p{18.5em} p{20.4em} ccc}
\toprule
\multirow{2}{18.5em}{\centering\arraybackslash\textbf{Original Text}} & \multirow{2}{20.4em}{\centering\arraybackslash\textbf{Split Text}} & \multirow{2}{3em}{\textbf{\textsc{Wiki}}} & \multicolumn{2}{c}{\textbf{\textsc{BiSECT}}}\\ 
& & & \textbf{en} & \textbf{de}\\
\midrule

\multicolumn{2}{c}{{\textbf{\textit{High-Quality Split and Rephrase pairs}}}} & \textbf{73\%} & \textbf{85\%} & \textbf{77\%}\\ \cmidrule{1-5}
\multirow{3}{18.5em}{An additional advantage is that a shorter ramp can be used, thereby reducing weight and improving the rear view of the driver. (de$\rightarrow$en)} & \multirow{3}{20.4em}{Another advantage is that a shorter ramp can be used. $\|$
This saves weight and improves the look of the rear of the vehicle.} &  \\
& & & & \\
& & \multicolumn{3}{c}{\textit{Perfect pairs}}\\
\cmidrule{1-2}
\multirow{4}{18.5em}{Bitte geben Sie hier Ihre E-Mail-Adresse ein und wir senden Ihnen anschlieend einen Link zu, mit dem Sie Ihr Passwort zurucksetzen konnen. (en$\rightarrow$de)} & \multirow{4}{20.4em}{Bitte geben Sie unten Ihre E-Mail-Adresse ein. $\|$ Wir senden Ihnen einen Link per E-Mail, mit dem Sie ein neues Passwort erstellen konnen.}  & 
\multirow{2}{*}{51\%} & \multirow{2}{*}{63\%} & \multirow{2}{*}{53\%} \\ 
& & & & \\
& & & & \\
& & & & \\
\cmidrule{1-5}

\multirow{4}{18.5em}{Its many \textcolor{DarkYellow}{\hlyellow{novel features}} ensure that \textcolor{DarkYellow}{\hlyellow{it}} is easy to use correctly, making it suitable for all patients regardless of disease severity, in the elderly and for children. (de$\rightarrow$en)} & \multirow{4}{20.4em}{Its numerous \textcolor{DarkYellow}{\hlyellow{control mechanisms}} ensure that \textcolor{DarkYellow}{\hlyellow{the Novolizer}} is easy to use correctly. $\|$ This makes it suitable for all patients regardless of the severity of the disease, for older patients and for children.}  & \multicolumn{3}{c}{\textcolor{DarkYellow}{\textit{Unsupported Details}}}  \\ 
& & & & \\
& & 21\% & 13\% & 18\% \\
& & & & \\ \cmidrule{1-5}

\multirow{3}{18.5em}{Every day, \textcolor{DarkYellow}{\hlyellow{pedestrians}} take risks \textcolor{DarkYellow}{\hlyellow{by working near mobile machinery}} and every day, accidents cost businesses dearly. (fr$\rightarrow$en)} & \multirow{3}{20.4em}{Every day, \textcolor{DarkYellow}{\hlyellow{men}} take risks \textcolor{DarkYellow}{\hlyellow{with machines}}. $\|$ And every day accidents cost businesses dearly.} & \multicolumn{3}{c}{\textcolor{DarkYellow}{\textit{Deleted Details}}}\\ 
& & \multirow{2}{*}{1\%} & \multirow{2}{*}{9\%} & \multirow{2}{*}{6\%}\\
& & & & \\ \midrule

\multicolumn{2}{c}{\textbf{\textit{Pairs with significant errors}}} & \textbf{27\%} & \textbf{15\%} & \textbf{23\%} \\
\cmidrule{1-5}

\multirow{4}{18.5em}{A little after the issue of Tosattis book, Rizzoli published another volume on Fatima, this time a book-interview with Cardinal Bertone, edited by Vatican expert Giuseppe De Carli. (de$\rightarrow$en)} & \multirow{4}{20.4em}{Shortly after the publication of Tosatti's book, the Italian publisher Rizzoli published another book on Fatima. $\|$ \textcolor{DarkRed}{\hlred{An interview book with Cardinal Bertone, edited by the Vaticanist Giuseppe De Carli.}}} & \multicolumn{3}{c}{\textcolor{DarkRed}{\textit{Disfluencies}}}  \\
& & & & \\
& & 10\% & 5\% & 12\% \\
& & & & \\ \cmidrule{1-5}
\multirow{4}{18.5em}{The children concoct many plans to lure Boo Radley out of his house for a few summers until Atticus \textcolor{DarkRed}{\hlred{make not true out}}, and they become ``engaged." (WikiSplit)} & \multirow{4}{20.4em}{The children concoct many plans to lure Boo Radley out of his house for a few summers until Atticus makes them stop. $\|$ \textcolor{DarkRed}{\hlred{Dill promises to marry Scout,}} and they become ``engaged."} & & & \\
& & & & \\
& & & & \\
& & \multicolumn{3}{c}{\textcolor{DarkRed}{\textit{Multiple Errors}}} \\
\cmidrule{1-2} 
\multirow{6}{18.5em}{Dann setzt unser Destillateurmeister die Brennblase in Gang und destilliert unter den Augen der Teilnehmer einen Berlin Dry Gin, der naturlich am Ende der Veranstaltung verkostet werdet kann. (en$\rightarrow$de)} & \multirow{6}{20.4em}{Distiller legt den noch in Bewegung in Bewegung und Destillern unter den Augen der Teilnehmer ein Berliner trockener Gin, der naturlich am Ende der Veranstaltung geschmeckt werden kann. $\|$ \textcolor{DarkRed}{\hlred{Und wahrend die noch Blasen, tauchen die Teilnehmer in die Welt des Gin ein.}}} & 17\% & 10\% & 11\%
\\
& & & & \\
& & & & \\
& & & & \\
& & & & \\
& & & & \\
\bottomrule
\end{tabular}
\setlength{\belowcaptionskip}{-10pt}
\caption{Examples of high-quality and noisy sentence splits in the \textsc{BiSECT} corpus. Some examples have \textcolor{DarkYellow}{\hlyellow{minor adequacy/fluency issues}} (not uncommon in most existing monolingual parallel corpora) and are still usable, while a small portion (15\%) contain more \textcolor{DarkRed}{\hlred{significant errors}}. Prevalence of each category is calculated based on 100 manually inspected pairs from \textsc{\textbf{Wiki}Split} \cite{botha2018learning} and English/German \textbf{\textsc{BiSECT}} (our work).} 
\label{tab:data_quality_check}
\vspace{2mm}
\end{table*}

\begin{table*}[ht!]
\setlength{\tabcolsep}{4pt}
\small
\centering
\begin{tabular}{p{18.5em} p{20.4em} cc}
\toprule
\centering\arraybackslash\textbf{Original Text} & \centering\arraybackslash\textbf{Split Text} & \textbf{\textsc{Wiki}} & \textbf{\textsc{BiSECT}} \\ \midrule
\multicolumn{2}{c}{{\textbf{\textit{Direct Insertion}}}} & \textbf{33\%} & \textbf{40\%} \\ \cmidrule{1-4}
\multirow{3}{18.5em}{Gaal the son of Ebed came with his brothers, and went over to Shechem\textcolor{DarkBlue}{\hlblue{; and}} the men of Shechem put their trust in him. (fr$\rightarrow$en)} & \multirow{3}{20.4em}{Gaal the son of Ebed came with his brethren, and they passed over to Shechem\textcolor{DarkGreen}{\hlgreen{.}} $\|$ The people of Shechem trusted him.} & \multicolumn{2}{c}{\textcolor{DarkBlue}{\textit{Colon/Semicolon}}} \\
& & \multirow{2}{*}{\centering 15\%} & \multirow{2}{*}{\centering 18\%} \\
& & & \\ \cmidrule{1-4}

\multirow{3}{18.5em}{When I play a MIDI file on my desktop, the sound quality is rich and clear\textcolor{DarkBlue}{\hlblue{, but when}} I play the same file on a laptop, it's not so great! (fr$\rightarrow$en)} & \multirow{3}{20.4em}{When I play MIDI files on my table extension the sound quality is excellent\textcolor{DarkGreen}{\hlgreen{.}} $\|$ \textcolor{DarkGreen}{\hlgreen{If}} I play them on my portable sound is no longer very good.} & \multicolumn{2}{c}{\textcolor{DarkBlue}{\textit{Conjunction}}} \\
& & \multicolumn{2}{c}{\textcolor{DarkBlue}{\textit{with subject}}} \\
& & \multirow{1}{*}{\centering 18\%} & \multirow{1}{*}{\centering 22\%} \\
& & & \\ \midrule

\multicolumn{2}{c}{\textbf{\textit{Changes Near Split}}} & \textbf{66\%} & \textbf{49\%}\\ \cmidrule{1-4}
\multirow{4}{18.5em}{The virus is carried and passed to others through blood or sexual contact \textcolor{DarkBlue}{\hlblue{and}} can cause liver inflammation, fibrosis, cirrhosis and cancer. (de$\rightarrow$en)} & \multirow{4}{20.4em}{The virus is transmitted to other people through blood or sexual contact\textcolor{DarkGreen}{\hlgreen{.}} $\|$ \textcolor{DarkGreen}{\hlgreen{It}} can cause liver inflammation, fibrosis, cirrhosis, and cancer.} & \multicolumn{2}{c}{\textcolor{DarkBlue}{\textit{Conjunction}}} \\
& & \multicolumn{2}{c}{\textcolor{DarkBlue}{\textit{without subject}}} \\
& & \multirow{2}{*}{\centering 18\%} & \multirow{2}{*}{\centering 13\%} \\
& & & \\ \cmidrule{1-4}

\multirow{3}{18.5em}{An additional advantage is that a shorter ramp can be used\textcolor{DarkBlue}{\hlblue{, thereby reducing}} weight and improving the rear view of the driver. (de$\rightarrow$en)} & \multirow{3}{20.4em}{Another advantage is that a shorter ramp can be used \textcolor{DarkGreen}{\hlgreen{.}} $\|$ \textcolor{DarkGreen}{\hlgreen{This saves}} weight and improves the look of the rear of the vehicle.} & \multicolumn{2}{c}{\textcolor{DarkBlue}{\textit{Gerund}}} \\
& & \multirow{2}{*}{\centering 7\%} & \multirow{2}{*}{\centering 10\%} \\
& & & \\ \cmidrule{1-4}

\multirow{3}{18.5em}{For the fur edge I choose the smudge tool with a dissolved brush and paint in the mask along the black edge \textcolor{DarkBlue}{\hlblue{to get}} a smooth transition. (de$\rightarrow$en)} & \multirow{3}{20.4em}{For the fur edge, I choose the tool with speckled brush tip and drag on the black edge in the mask\textcolor{DarkGreen}{\hlgreen{.}} $\|$ \textcolor{DarkGreen}{\hlgreen{This creates}} a transition \textcolor{DarkGreen}{\hlgreen{to the background.}}} &
\multicolumn{2}{c}{\textcolor{DarkBlue}{\textit{Preposition /}}} \\
& & \multicolumn{2}{c}{\textcolor{DarkBlue}{\textit{Subordinate clause}}} \\
& & \multirow{2}{*}{\centering 17\%} & \multirow{2}{*}{\centering 9\%} \\
& & & \\ \cmidrule{1-4}

\multirow{4}{18.5em}{Over 3500 people visit the Centre every year \textcolor{DarkBlue}{\hlblue{where}} they are greeted by volunteers who show them around the study room and tell them about the collection. (fr$\rightarrow$en)} & \multirow{4}{20.4em}{Each year, more than 3,500 people visit the Center\textcolor{DarkGreen}{\hlgreen{.}} $\|$ They are greeted by volunteers who show them the study room and introduce them to the collection.} & \multicolumn{2}{c}{\textcolor{DarkBlue}{\textit{Concluding}}} \\
& & \multicolumn{2}{c}{\textcolor{DarkBlue}{\textit{Relative Clause}}} \\
& & \multirow{2}{*}{\centering 24\%} & \multirow{2}{*}{\centering 17\%} \\
& & & \\ \midrule

\multicolumn{2}{c}{\textbf{\textit{Changes Across Sentence}}} & \textbf{1\%} & \textbf{11\% $\uparrow\uparrow$} \\ \cmidrule{1-4}
\multirow{3}{18.5em}{\textcolor{DarkBlue}{\hlblue{Because}} these cities, settlements and regions were constructed for not hundred years\textcolor{DarkBlue}{\hlblue{, but}} for centuries. (fr$\rightarrow$en)} & \multirow{3}{20.4em}{All these towns, these localities were not built in a hundred years\textcolor{DarkGreen}{\hlgreen{.}} $\|$ \textcolor{DarkGreen}{\hlgreen{They were created}} over the centuries.} & \multicolumn{2}{c}{\textcolor{DarkBlue}{\textit{Preceding}}} \\
& & \multicolumn{2}{c}{\textcolor{DarkBlue}{\textit{Relative Clause}}} \\
& & \multirow{1}{*}{\centering 1\%} & \multirow{1}{*}{\centering 11\%} \\ \bottomrule
\end{tabular}
\setlength{\belowcaptionskip}{-10pt}
\caption{Categories in Split and Rephrase tasks with examples and frequency observed in the \textsc{\textbf{Wiki}Split} \cite{botha2018learning} and the English \textsc{BiSECT} (our work) corpora. Categories grouped under \textbf{\textit{Direct Insertion}} require extremely minor changes in order to split the sentence; categories under \textbf{\textit{Changes Near Split}} require some minor modifications around the source of the split; and categories under \textbf{\textit{Changes Across Sentence}} require more major changes across the original sentence. Statistics are based on manual inspection of 100 examples from each corpus.}
\label{tab:split_categories}
\end{table*} 

\subsection{Comparison to Existing Corpora}
\label{sec:manual_quality}

\paragraph{Corpus Statistics.} Besides corpus size, we are interested in the amount of rephrasing (indicated by \%new) and the syntactic complexity of sentences (approximated by length). In Table \ref{tab:corpus_statistics}, we compare  \textsc{BiSECT} with previous split and rephrase corpora, including \textsc{WikiSplit}  \cite{botha2018learning}, \textsc{WebSplit} \cite{narayan2017split,aharoni2018split}, HSplit-Wiki \cite{sulem2018bleu}, Contract and Wiki-BM \cite{zhang2020small}. 
\textsc{BiSECT} is comparable in size with \textsc{WikiSplit}, while importantly containing longer aligned sentence pairs and a higher \%new score, indicating that \textsc{BiSECT} contains more complex pairs with significantly more rephrasing (see also examples in Tables \ref{tab:data_quality_check} and \ref{tab:split_categories}).  

\paragraph{Manual Quality Assessment.} While BiSECT does not suffer from   meaning-altering edits like \textsc{WikiSplit} does, a potential concern is the error induced from translating a foreign text to English. Thus, we perform a manual assessment of corpus quality by comparing 100 randomly selected pairs from both \textsc{BiSECT} and \textsc{WikiSplit} corpora. We categorize each example $(l, s)$ into two groups: (1) \textit{\textbf{high-quality pairs}}, where both $l$ and $s$ are grammatical, $l$ consists of exactly one sentence, and $s$ contains exactly two sentences; and (2) \textit{\textbf{significant errors}}, where the pair contains drastic errors impacting its usability. Table \ref{tab:data_quality_check} shows the results of the manual inspection. When compared with \textsc{WikiSplit}, \textsc{BiSECT} contains significantly more high-quality pairs, while containing fewer pairs with significant errors. Pairs containing unsupported and deleted details are comparable across corpora, though \textsc{WikiSplit}  skews more towards adding unsupported information, which is consistent with previous work \cite{zhang2020small}. 

Moreover, we take 100 random samples from the German \textsc{BiSECT} corpus and perform manual inspection. We chose 
German because translating to/from German is notoriously challenging for translation systems \cite{twain1880awful,collins2005clause}. As shown in Table \ref{tab:data_quality_check}, German \textsc{BiSECT} still contains 77\% high-quality pairs. 

\subsection{Categorization for Split and Rephrase}

One aspect of the Split and Rephrase task that has received little attention, outside of \newcite{zhang2020small}, is the amount of rephrasing that occurs in each instance, and more specifically the syntactic patterns involved in this rephrasing. Unlike more open-ended language generation tasks, the structural paraphrasing involved in Split and Rephrase is likely to be relatively consistent across domains, thus identifying these patterns is a critical step towards further improvement of neural-based approaches. In this work, we consider three major categories, and break down each of these further into more specific syntactic patterns. The categories are derived from the entire dataset, spanning the domains of web, newswire, medical and legal text, and others.

The first group involves \textbf{\textit{Direct Insertion}}, when a long sentence $l$ contains two independent clauses, and requires only minor changes in order to make a fluent and meaning-preserving split $s$. Within this category, we identify two sub-categories: \textit{Colon/Semicolon}, which occurs when the clauses are connected by a colon or semicolon; and \textit{Conjunction with subject}, where the clauses are connected by a conjunction, and the second clause contains an explicit subject. The second group involves \textbf{\textit{Changes near Split}}, when $l$ contains one independent and one dependent clause, but modifications are restricted to the region where $l$ is split. Within this category, we identify four sub-categories: instances containing a \textit{conjunction without subject}, which involves two clauses connected by a conjunction, but the second clause does not have an explicit subject; instances that contain a \textit{gerund}, followed by an adjectival clause, adverbial clause, or prepositional phrase; instances that involve an explicit \textit{subordinate clause}; and instance that contain a \textit{concluding relative clause}. Finally, the third major group involves \textbf{\textit{Changes across Sentences}}, where major changes are required throughout $l$ in order to create a fluent split $s$. 
The main subcategory within this group involves a \textit{preceding relative clause}, followed by a comma.

Table \ref{tab:split_categories} presents the examples and prevalence of each category in \textsc{WikiSplit} and \textsc{BiSECT}, computed using a manual inspection of 100 random examples from each corpus. \textsc{BiSECT} contains significantly more instances that require changes across the sentence to form a high-quality split. 
To assess the relative difficulty of these categories, we analyze the quality of sentence splits generated by \textbf{DisSim} \cite{niklaus-etal-2019-transforming}, a rule-based sentence splitter, on these 200 selected examples. DisSim splits the source sentence recursively using 35 hand-crafted rules based on a syntactic parse tree. DisSim produces disfluent sentence splits 34\% of the time, and performs no splitting 9\% of the time. For the \textit{Changes near Split} and \textit{Changes Across Sentence} categories, the number of erroneous splits increases to 55\% and 63\%, respectively. Although rules correctly identify the location of sentence splits, they fail to effectively modify sentences requiring more expansive rephrasing.

\section{Our Model}

The \textsc{BiSECT} corpus contains a significant amount of paraphrasing along with sentence splitting, and models trained on \textsc{BiSECT} tend to alter the lexical choices made in the input sentence.  Although this is desirable in some situations, like for the task of sentence simplification, sometimes it can alter the meaning of the input sentence. We propose a novel model that allows finer-grained control over what parts of the sentence are changed. Our approach leverages the sentence split categories described in \S 3.3 to identify the split-based edits and incorporates them into a customized loss function as distantly supervised labels. This section describes the base model and its variant that adapts a high paraphrasing \textsc{BiSECT} corpus to a sentence splitting task with minimal rephrasing. 


 \subsection{Base Model}

Our base model is a \textbf{BERT-Initialized Transformer}  \cite{rothe2020leveraging}, a state-of-the-art model for Split and Rephrase. The encoder and decoder follow the BERT$_{base}$ architecture, with the encoder initialized with the same checkpoint. The base model is trained using standard cross-entropy loss. During training, the split sentences in the reference are separated by a separator token $[SEP]$.

\subsection{Adaptive Loss using Distant Supervision}

The base model treats all the sentence splitting categories (Table \ref{tab:split_categories}) similarly even though the edits necessary to split the sentence vary across the categories. 
We utilize heuristics and linguistic rules to categorize each source-target sentence pair and extract required edits based on the category. Finally, we train the base model on these classification and edit labels to guide the model to perform appropriate edits for each category.



\paragraph{Classification and Edit Labels.} Given the source $\mathbf{x}=(x_1, x_2, \ldots x_N)$ and target $\mathbf{y}=(y_1, y_2, \ldots y_N)$, we assign a sentence category label $l \in \{``\text{\textit{Direct Insertion}}"$,$``\text{\textit{Changes Near Split}}"$ ,$``\text{\textit{Changes Across Sentence}}"\}$ to the training pair, and a binary label $\delta_i$ to each position indicating whether the word is modified from the input. Here, $\boldsymbol{\delta}=(\delta_1, \delta_2, \ldots \delta_N)$ represent the edit labels and $\delta_i = 1$ represents the necessary changes to split the sentence that cannot be copied from $\mathbf{x}$. We ensure that $\mathbf{x}$ and $\mathbf{y}$ are of the same length using padding around the split. The split position for $\mathbf{y}$ corresponds to the position of the $[SEP]$ token. For $\mathbf{x}$, we extract the lexical differences between $\mathbf{x}$ and $\mathbf{y}$ using an edit distance algorithm\footnote{\url{https://pypi.org/project/simplediff/}} and label the edit in $\mathbf{x}$ close to the $[SEP]$ token in $\mathbf{y}$ as the split position.  Finally, we pad the sequences before and after the split positions so that they are of equal length. We provide an example in Appendix \ref{app:model_design}.

We extract $l$ for each pair using the following rules: (1) If the first level of the parse tree of $\mathbf{x}$ contains the pattern ``$S\;CC\;S$'', $\mathbf{x}$ contains a colon/semicolon, or the lexical differences between $\mathbf{x}$ and $\mathbf{y}$ contain only the split, then we label the pair as \textit{Direct Insertion}. Once again, we extract lexical differences using an edit distance algorithm. (2) If the first level parse tree of $\mathbf{x}$ contains the pattern ``$S\;NP\;VP$'' or ``$SBAR\;NP\;VP$'', then we label the pair as \textit{Changes across sentence}. (3) If the first level of the parse tree contains ``$VP\;CC\;VP$'' or at least 5 words at the beginning and end of the sentence are copied from the source, then we categorize the pair as \textit{Changes near split}. (4) We label the rest as \textit{Changes across sentence}. In case of multiple potential splits, we choose the split whose lengths 
is closest to that of the reference.

After extracting $l$, we construct $\boldsymbol{\delta}$ using the lexical overlap between $\mathbf{x}$ and $\mathbf{y}$. For \textit{Direct Insertion}, we set the $\delta_i$ corresponding to the split position and its adjacent positions to $1$ to capture the punctuation and capitalization. For \textit{Changes near split}, we construct a variable length window around the split position to facilitate the addition of the new words and set the $\delta_i$ in the window to $1$. To construct this window, we scan the sequence on each side of the split position  until the position where at least 3 consecutive positions are copied from $\mathbf{x}$ to $\mathbf{y}$.  Finally, we set $\boldsymbol{\delta}$ to a one vector for \textit{Changes Across Sentence}, as the changes cannot be localized. Our manual inspection of 100 training pairs from the \textsc{BiSECT} training set showed that the rules correctly classified 83\% of the pairs.

\paragraph{Distant Supervision.} 
As $l$ depends on the reference and cannot be used during inference, we introduce a multi-class classification task distantly supervised by $l$. We train our model in a multi-task learning setting to predict $l$ and perform generation. The classifier predicts the probability that $\mathbf{x}$ belongs to a split category using the encoder representation of the $[CLS]$ token prepended to the input by the BERT encoder. The classifier contains a linear layer with a $softmax$ activation function. 




While $l$ represents the sentence category, $\delta$ captures split-related edits.
To ensure our model learns only split-based edits, we combine $\mathbf{x}$ and $\mathbf{y}$ in our decoder generation loss ($L_{seq}$) using $\delta$ as follows: 

\vspace{-5mm}
\begin{equation*}
\begin{split}
L_{seq}= &  \frac{1}{m} \sum_{i=1}^{m} (1 - \delta_i) P(x_i | \hat{y}_{<i} ) + \delta_i P(y_i | \hat{y}_{<i} )\\ 
\hat{y}_i= & (1 - \delta_i) x_i + \delta_i y_i \\
 \delta_i=  &
\begin{cases}
    0, & \text{if } x_i \text{ is copied}\\
    1, & \text{otherwise}
\end{cases} \\
\end{split}
\end{equation*}

\noindent where $m$ is the number of training examples and $\hat{y}_{<i}$ represents the mixture of of $\mathbf{x}$ and $\mathbf{y}$ histories. In other words, our model only learns the edits where $\delta_i = 1$ and copies from the source sentence for the rest of the positions. Finally, we jointly train the classifier and the Transformer using the cross entropy loss and our custom split-focused loss. We provide model and training details in Appendix \ref{app:implementation}.

\begin{table*}[ht!]
\small
\centering
\setlength{\tabcolsep}{4pt}
\renewcommand{\arraystretch}{0.95}
\resizebox{0.98\textwidth}{!}{%
\begin{tabular}{l|cccc|c|cccc|cc}
\toprule
 \textbf{Models w/ Training Data}  & \textbf{SARI} & \textbf{add} & \textbf{keep}  & \textbf{del} & \textbf{BScore} &\textbf{FK} & \textbf{BLEU} & \textbf{SLen} & \textbf{OLen}  & \textbf{sBLEU} &  \textbf{\%new}   \\ \midrule 
\multicolumn{11}{l}{\textbf{\textsc{BiSECT} test set} } \\
\cmidrule{1-12}
Source & 20.1 & 0.0 & 60.3 & 0.0 & 84.6 & 19.2 & 43.5 & 35.2 & 39.9 & 100.0 & 0.0\\
DisSim & 40.0 & 2.4 & 55.2 & 62.4 & 76.9 & 11.2 & 30.0 & 12.4 & 40.6 & \textbf{61.4} & \textbf{18.7} \\
Copy512 w/ \textsc{Wiki} & 46.7 & 4.0 & 61.6 & 74.6 & 84.1 & 13.0 & 43.0 & 19.4 & 39.7 & 88.5 & 4.5 \\
Copy512 w/ \textsc{BiSECT} & 52.7 & 10.6 & 64.8 & \textbf{82.8} & 85.3 & \textbf{12.6} & \textbf{46.3} & 18.5 & 39.2 & 81.5 & 6.7 \\
Transformer w/ \textsc{Wiki} & 49.3 & 6.9 & 62.8 & 78.2 & 84.5 & 12.4 & 43.1 & \textbf{19.3} & \textbf{41.0} & 81.8 & 9.3 \\
Transformer w/ \textsc{BiSECT}  & \textbf{55.5} & \textbf{18.3} & \textbf{66.9} & 81.4 & \textbf{85.6} & 12.1 & 45.8 & 19.0 & 40.7 & 63.9 & 16.6 \\
Transformer$_{control}$ w/ \textsc{BiSECT} & 49.0 & 7.9 & 62.6 & 76.1 & 84.8 & \textbf{12.6} & 42.9 & 19.8 & 40.9 & 79.7 & 10.5 \\
Transformer$_{control}$ w/ \textsc{BiSECT+Wiki} & 47.7 & 6.0 & 62.1 & 75.0 & 84.3 & 12.9 & 43.5 & \textbf{19.1} & 39.7 & 85.5 & 6.0 \\
\cmidrule{1-12}
Reference & 94.3 & 88.8 & 97.9 & 96.1 & 100.0 & 12.5 & 100.0 & 19.2 & 41.5  & 40.4 & 32.0  \\
\midrule
\multicolumn{11}{l}{\textbf{\textsc{HSplit-Wiki} test set} } \\
\cmidrule{1-12}

Source & 30.5 & 0.0 & \textbf{91.4} & 0.0 & \textbf{ 97.1}{ } & 12.6 & \textbf{83.0} & 22.4 & 22.6 & 100.0 & 0.0 \\
DisSim & 38.0 & 5.0 & 79.3 & 29.6 & 87.7 & 8.9 & 52.5 & 10.5 & 23.6 & 62.8 & 17.1 \\
Copy512 w/ \textsc{Wiki}$^\dagger$ &  47.2 & 13.0 & 87.9 & 40.8 & 93.3 & \textbf{8.4} & 68.2 & 12.3 & 24.7 & 71.2 & 17.0 \\
Copy512 w/ \textsc{BiSECT} & 47.4 & 13.6 & 87.4 & 40.7 & 92.3 & 8.3 & 69.0 & 12.0 & 23.6 & 72.2 & 14.3 \\
Transformer w/ \textsc{Wiki}$^\dagger$ & 49.5 & 14.9 & 88.4 & 45.2 & 95.3 & 7.8 & 69.2 & 12.0 & \textbf{24.8} & 73.1 & 15.8 \\
Transformer w/ \textsc{BiSECT}  & 45.7 & 17.7 & 80.2 & 39.1 & 92.0 & 7.8 & 57.8 & \textbf{12.7} & 26.2 & 57.0 & 26.2 \\
Transformer$_{control}$ w/ \textsc{BiSECT}  & 47.2 & 13.3 & 87.2 & 41.1 & 94.1 & 7.9 & 67.2 & 12.3 & 24.9 & 70.9 & 17.6 \\
Transformer$_{control}$ w/ 
\textsc{BiSECT+Wiki} & 
\textbf{52.0} & \textbf{15.7} & 90.4 & \textbf{50.0} & 95.4 & 8.3 & 74.0 & 11.9 & 23.9 & \textbf{78.2} & \textbf{11.9} \\
\cmidrule{1-12}
Reference & 60.1 & 33.0 & 94.1 & 53.2 & 100.0 & 8.4 & 100.0 & 12.6 & 24.3 & 81.8 & 10.6 \\
\bottomrule
\end{tabular}}
\caption{Automatic and human evaluation results on \textsc{BiSECT} and \textsc{HSplit-Wiki} test sets. We report \textbf{SARI} and its three edit scores, namely precision for delete (\textbf{del}) and F1 scores for \textbf{add} and \textbf{keep} operations. We also report BERTScore (\textbf{BScore}), FKGL (\textbf{FK}), corpus-level BLEU (\textbf{BLEU}), average number of words in a sentence  (\textbf{SLen}), average number of words in the output (\textbf{OLen}), self-BLEU (\textbf{sBLEU}), and average percentage of new words added to the output (\textbf{\%new}). \textbf{Bold} typeface denotes the best performances (i.e., closest to the reference). $^\dagger$ These models have a natural advantage on the i.i.d. sampled Wiki-based \textsc{HSplit} test set, as they are trained on \textsc{WikiSplit} data. In contrast, 
 the train and test data in \textsc{BiSECT} are not i.i.d. sampled and from different sources (Table \ref{tab:corpus_statistics}).} 
\label{table:automatic_eval}
\end{table*}

\section{Experiments and Results}

In this section, we compare different split and rephrase models trained on our new \textsc{BiSECT} corpus. We also conduct a carefully designed human evaluation as automatic metrics are not totally reliable. Our model trained on \textsc{BiSECT} establishes a new start-of-the-art for the task.

\setlength{\tabcolsep}{2pt}
\begin{table}[t!]
\small
\centering
\begin{tabular}{l|cc}
\toprule
\textbf{Model} & \textsc{\textbf{BiSECT}} &\textsc{\textbf{HSplit-Wiki}} \\
\midrule
Random & 13.2 &  11.9 \\
Transformer w/ \textsc{Wiki} & 88.2 & 88.1 \\
Transformer w/ \textsc{BiSECT} & 93.8 & \textbf{92.0} \\
Transformer$_{control}$ w/ B  & \textbf{94.8} & 84.5 \\
Transformer$_{control}$ w/ \textsc{B+W} & 89.2 & 88.5  \\
Reference & 95.0 & 96.8 \\
\bottomrule
\end{tabular}
\setlength{\belowcaptionskip}{-6pt}
\caption{Human evaluation of the overall sentence splitting quality (rating on 0-100 scale) on 100 examples from the \textsc{BiSECT} and \textsc{HSplit-Wiki} test sets. \textbf{\textsc{B}} and \textbf{\textsc{W}} represent \textsc{BiSECT} and \textsc{WikiSplit} respectively.} 
\label{table:human_eval}
\end{table}

\subsection{Data and Baselines}

We train the models on \textsc{BiSECT} and  \textsc{WikiSplit} corpora. For evaluation, we select the \textsc{BiSECT} and \textsc{HSplit-Wiki} \cite{sulem2018bleu} test sets to represent splitting with a high degree and minimal of rephrasing respectively. \textsc{HSplit-Wiki} is a human annotated dataset with 359 complex sentences and 4 references for each complex sentence.  Following previous work \cite{botha2018learning,zhang2020small}, we do not use \textsc{WikiSplit} for evaluation, because this corpus was constructed explicitly to be used only as training data, as it contains inherent noise and biases.  While \textsc{BiSECT} contains 928,440/9,079 train and dev pairs, \textsc{WikiSplit} contains 989,944/5,000 train and dev pairs. Note that we constructed  \textsc{BiSECT} test set by manually selecting 583 high-quality sentence splits from 1000 random source-target pairs from \textsc{EMEA} and \textsc{JRC-Acquis} corpora. 

We compare our approach with \textbf{Copy512} \cite{aharoni2018split}, a state-of-the-art model consisting of an attention-based LSTM encoder-decoder with a copy mechanism \cite{see2017get}. We use our base model trained on \textsc{WikiSplit} \cite{rothe2020leveraging} as another state-of-the-art baseline.

\subsection{Automatic Evaluation}

Existing automatic metrics, such as BLEU \cite{papineni2002bleu} and SAMSA \cite{sulem2018bleu}, are not optimal for the Split and Rephrase task as they rely on lexical overlap between the output and the target (or source) and underestimate the splitting capability of the models that rephrase often. We focus on BERTScore \cite{zhang2020bertscore} and SARI \cite{xu2016optimizing}. BERTScore \cite{zhang2020bertscore} captures meaning preservation and fluency well \cite{scialom2021rethinking}. SARI can provide three separate F1/precision scores that explicitly measure the correctness of inserted, kept and deleted n-grams when compared to both the source and the target. We use an extended version of SARI that considers lexical paraphrases of the reference. An n-gram from the output is considered correct if the given n-gram or its paraphrase from PPDB \cite{pavlick2015ppdb} occurs in the reference, using the PPDB-L version. Without this change, the original SARI also tends to underestimate rephrasing.



Table \ref{table:automatic_eval} shows that our models trained on \textsc{BiSECT} outperform their equivalents trained on \textsc{WikiSplit} in terms of SARI and BERTScore. Note that the models trained on \textsc{Wikisplit} have an advantage over \textsc{HSplit-Wiki} test set because they belong to the same domain. Models trained on \textsc{BiSECT} do not have a similar advantage on \textsc{BiSECT} test set because it belongs to a different domain than the training data. When compared to the base model (Transformer w/ \textsc{BiSECT}), our model (Transformer$_{control}$ w/ \textsc{BiSECT}) shows higher self-BLEU and lower percentage of new words, indicating that it performs less rephrasing by focusing on split-based edits.


\begin{table*}[h!]
\setlength{\tabcolsep}{2pt}
\small
\centering
\begin{tabular}{p{2.5cm}|p{12.5cm}}
\toprule
\textbf{Model w/ Data} & \textbf{System Outputs} \\
\midrule
Source &  \hlblue{Having determined, after consulting} the Advisory Committee  \hlblue{that} sufficient evidence existed for the initiation of a partial interim review\hlblue{, the} Commission published a notice in the Official Journal of the European Communities and commenced an investigation. \\
Transformer w/ \textsc{Wiki} &  \hlred{Having determined, after consulting} the Advisory Committee\hlred{, that} sufficient evidence existed for the initiation of a partial interim review\hlgreen{. The} Commission published a notice in the Official Journal of the European Communities and commenced an investigation. \\
Transformer w/ \textsc{BiSECT} &  \hlgreen{After consulting} the Advisory Committee, \hlgreen{the Commission determined that there was} sufficient evidence for the initiation of a partial interim review\hlgreen{. The} Commission issued a notice in the Official Journal of the European Communities and began an investigation. \\
Transformer$_{control}$ w/ \textsc{BiSECT} &  \hlgreen{After consulting} the Advisory Committee\hlgreen{, there was} sufficient evidence for the initiation of a partial interim review\hlgreen{. The} Commission published a notice in the Official Journal of the European Communities and initiated an investigation. \\
\bottomrule
\end{tabular}
\caption{Examples of system outputs from the \textsc{BiSECT} test set. Here, the source sentence belongs to the category ``\textbf{Changes Across Sentence}''. \hlblue{Blue} marks the location of the required edits in the source sentence. \hlgreen{Green} indicates good edits and \hlred{red} indicates errors.}
\label{tab:sample_outputs}
\end{table*}

\subsection{Human Evaluation}
\label{sec:human_eval}
We asked three annotators to rate the overall quality of the sentence splits generated by different models on a 0-100 point scale. 0 represents an erroneous split and 100 represents a perfect meaning-preserving split. Unlike the previous work that measures meaning preservation and fluency separately, we collected only one rating because it was difficult to distinguish between the grammatical and the meaning-changing errors. We modeled our evaluation after the WMT evaluation \cite{ws-2019-machine-translation-3} that also uses a similar setting. We evaluated on 100 random sentences from the \textsc{BiSECT} and \textsc{HSplit-Wiki} test sets. The annotators were university students trained using an instructional video and a qualification phase. To capture the annotation quality, we included a control output generated by randomly selecting a system output and replacing 4 to 8 words with random words. Our annotators gave low ratings ($<$20) to the control outputs, indicating that the ratings are reliable. We provide the annotation interface design in Appendix \ref{app:human_eval}.


\begin{figure}[bt]
\centering
\includegraphics[width=0.98\linewidth]{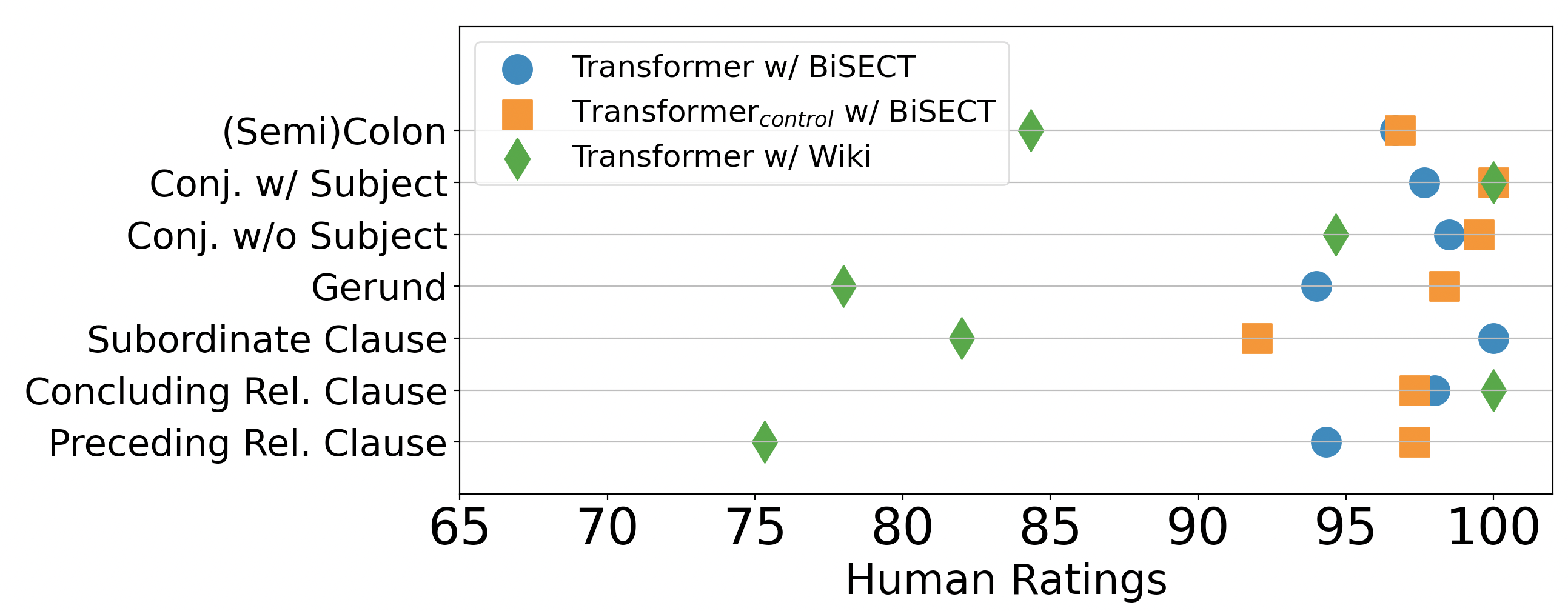}
\caption{Human ratings on 100 generated sentence splits from the \textsc{BiSECT} test set broken down by sentence split categories as described in Table \ref{tab:split_categories}.}
\label{fig:category_results}
\end{figure}

Table \ref{table:human_eval} shows that results on the entire \textsc{BiSECT} and \textsc{Wiki-HSplit} test sets. Figure \ref{fig:category_results} shows the results on different split categories of the \textsc{BiSECT} test set. The sentences splits generated by models trained on \textsc{BiSECT} are of better quality than the ones trained on \textsc{WikiSplit}. Our model with adaptive loss (Transformer$_{control}$ w/ \textsc{BiSECT}) performs  better than the base model (Transformer w/ \textsc{BiSECT}) in four of the seven split categories. The difference in quality is much more evident for the \textit{Preceding Relative Clause} category, as this requires changes across sentences. We provide an example in Table \ref{tab:sample_outputs}, as well as several more in Appendix \ref{app:system_outputs}.

\section{Conclusion}
In this work, we introduce \textsc{BiSECT}, a new corpus for the Split and Rephrase task in several languages. We create this by making use of bilingual parallel corpora, and translating instances of aligned split sentences. We show that the sentence splitting models trained on our new corpus generate fewer errors than their counterparts trained on the existing datasets. To further improve meaning preservation and diversity, we propose a novel approach that identifies split-related edits in a training pair using linguistic rules and trains the model solely on split-based edits. Our proposed approach trained on \textsc{BiSECT} outperforms existing systems in terms of both automatic and human evaluations. We plan to investigate and create better automatic evaluation metrics for future work. 

\section*{Acknowledgments}

We thank four anonymous reviewers for their helpful comments. We also thank Andrew Duffy, Manish Jois, Riley Kolb, Sounak Dey, Alyssa Hwang, Bryan Li, Veronica Qing Lyu and UPenn NETS 212 students for assisting through the human evaluation iterations. This research is supported in part by the NSF awards IIS-2055699, ODNI and IARPA via the BETTER program contract 19051600004, and the DARPA KAIROS Program (contract FA8750-19-2-1004). The views and conclusions contained herein are those of the authors and should not be interpreted as necessarily representing the official policies, either expressed or implied, of NSF, ODNI, IARPA, DARPA, or the U.S. Government. The U.S. Government is authorized to reproduce and distribute reprints for governmental purposes notwithstanding any copyright annotation therein.

\bibliographystyle{acl_natbib}
\bibliography{emnlp2021}

\newpage
\appendix

\clearpage

\section{Implementation and Training details}
\label{app:implementation}
We implemented the BERT-intialized Transformer using the Fairseq\footnote{https://github.com/pytorch/fairseq} toolkit. Here, the encoder and decoder follow BERT$_{base}$\footnote{https://github.com/google-research/bert} architecture. The encoder is also initialized with BERT$_{base}$ checkpoint and the decoder is randomly initialized. The sentence classifier is a feedforward network containing an inputer layer, one hidden layer with 1000 nodes, and an output layer with 3 nodes and \textit{softmax} activation. We used Adam optimizer \cite{Kingma2014} with a learning rate of 0.0001, linear learning rate warmup of 40k steps, and 100k training steps. We used a batch size of 64. We used BERT WordPiece tokenizer. During inference, we use beam-search of width 10 and ensure that the beam-search does not repeat trigrams.  We used the hyperparameters of the BERT-initialized Transformer described in \citet{rothe2020leveraging}. The model takes 10 hours to train on 1 NVIDIA GeForce GPU.

\onecolumn

\section{BiSECT Language Composition}
\label{app:bisect_composition}

\begin{table*}[h!]
\setlength{\tabcolsep}{4pt}
\centering
\small
\begin{tabular}{l|cccccccc}
 \toprule
   \textbf{Dataset} & \textbf{French} & \textbf{German} & \textbf{Spanish} & \textbf{Arabic} & \textbf{Dutch} & \textbf{Italian} & \textbf{Portuguese} & \textbf{Russian} \\
   \midrule
   \textsc{CCAligned} & 204K & -- & -- & -- & -- & -- & -- & --  \\
   \textsc{Europarl} & 57K & -- & -- & -- & -- & -- & -- & -- \\
   \textsc{Gigaword} & 132K & -- & -- & -- & -- & -- & -- & -- \\
   \textsc{Paracrawl} & 125K & 144K & 66K & -- & 31K & 26K & 14K & --\\
   \textsc{UN} & 5.7K & & 8K & 36K & -- & & -- & 6.7K \\
   \textsc{EMEA} & 1K & -- & -- & -- & -- & -- & -- & -- \\
   \textsc{JRC-Acquis} & 1K  & 3.7K & -- & -- & -- & -- & -- & -- \\
   \cmidrule{1-9}
   \textsc{Total} & 672K & 151K & 75K & 36K & 33K & 27K & 15K & 6.7K \\
\bottomrule
\end{tabular}
\caption{Composition of the \textsc{BiSECT} (English) corpus by pivoted language over bilingual parallel corpora.}
\label{tab:bisect_language_composition}
\end{table*}


\section{Examples from Different Corpora}
\label{app:corpora_examples}

\setlength{\tabcolsep}{4.0pt}
\begin{table*}[h!]
\scriptsize
\centering
\begin{tabular}{p{1.5cm}|p{14cm}}
  \toprule \textbf{Corpus} 
  &\textbf{Examples} \\

\midrule
\textsc{Wiki-Auto} & \textbf{Source:} Following the establishment of the Pembrokeshire Coast National Park in \hlblue{1952, Welsh naturalist and author Ronald Lockley surveyed} a route around the coast. \\
& \textbf{Reference:} The Pembrokeshire Coast National Park \hlgreen{was founded} in 1952\hlgreen{. After it was founded,} Ronald Lockley \hlgreen{did a survey for a path on the coastline.} \\
\cmidrule{1-2}
\textsc{Newsela-Auto} & \textbf{Source:} About 160,000 Girl Scouts participated in the program over the past \hlblue{year and were credited} with selling nearly 2.5 million boxes of cookies beyond those sold through traditional in-person methods. \\
& \textbf{Reference:} About 160,000 Girl Scouts \hlgreen{used} \hlred{Digital Cookie} \hlgreen{last year . They sold almost} 2.5 million boxes of cookies online.\\
\midrule
\textsc{HSplit} & \textbf{Source:} West Berlin had its own postal administration\hlblue{, separate} from West Germany's\hlblue{, which} issued its own postage stamps until 1990. \\
& \textbf{Reference:} West Berlin had its own postal administration\hlgreen{. It was} separate from West Germany's\hlgreen{. West Berlin} issued its own postage stamps until 1990. \\
\cmidrule{1-2}
\textsc{Contract} & \textbf{Source:} Except for Supplier's obligations and liability resulting from Section 10.0, Supplier Liability for Third Party Claims\hlblue{, Supplier's} liability for any and all claims will be limited to the amount of \$1,000,000 USD per occurrence\hlblue{, with} an aggregated limit of \$4,500,000 USD during the term of this Agreement . \\
& \textbf{Reference:} The following applies, not including the Supplier 's obligations and liability resulting from Section 10.0, Supplier Liability for Third Party Claims\hlgreen{. Supplier's} liability for any and all claims will be limited to the amount of \$1,000,000 USD per occurrence\hlgreen{. Additionally, there is} an aggregated limit of \$4,500,000 USD during the term of this Agreement .\\
\cmidrule{1-2}
\textsc{Wiki-BM} & \textbf{Source:} Together with James, she compiled crosswords \hlblue{for several} newspapers and magazines, including People\hlblue{, and it} was in 1978 \hlblue{that they launched} their own publishing company. \\
& \textbf{Reference:} Together with James, she compiled crosswords\hlgreen{. It} was for several newspapers and magazines, including People\hlgreen{. They} launched their own publishing company\hlgreen{. It} was in 1978.\\
\cmidrule{1-2}
\textsc{WebSplit v1.0} & Elliot See (\hlblue{born on} July 23, 1927 in Dallas \hlblue{and died} on February 28, 1966 in St Louis) \hlblue{was an American who graduated} from the University of Texas at Austin.\\
& Elliot See \hlgreen{attended} the University of Texas at Austin\hlgreen{. Elliot See}, \hlred{deceased}, was born in Dallas\hlgreen{. Elliot See} died on February 28, 1966, in St Louis\hlgreen{. Elliot See} was born on July 23, 1927\hlgreen{. Elliot See} is \hlgreen{a United States national}.\\
\cmidrule{1-2}
\textsc{WikiSplit} & In 2006, he and the Cavaliers negotiated a three-year, \$ 60 million contract \hlblue{extension instead of} the four year maximum as it allotted him the option of seeking a new contract worth more money as an unrestricted free agent following the 2010 season . \\
& In 2006, he and the Cavaliers negotiated a three-year, \$ 60 million contract extension\hlgreen{. This was} instead of the four year maximum \hlgreen{length} as it allotted \hlred{James} the option of seeking a new contract worth more money as an unrestricted free agent following the 2010 season .\\
\cmidrule{1-2}
\textsc{BiSECT} & Respondents felt that headsets compatible with hearing aids would greatly assist them in understanding what is being said\hlblue{, and added that} headsets in business class or first class on some aircraft are already compatible with hearing aids.\\
& Participants \hlgreen{indicated} that \hlgreen{the installation of} headsets \hlgreen{which are} compatible with hearing aids would 
\hlgreen{improve their ability to understand} what was being said \hlgreen{. It was mentioned} that headsets in the business or first class \hlgreen{portions of} some aircraft are already \hlgreen{hearing aid compatible}.\\
\bottomrule
\end{tabular}
\caption{Random examples of sentence pairs from the existing corpora. \hlblue{Blue} indicates the position of sentence splits in the source sentence.   \hlgreen{Green} indicates good edits, and \hlred{red} indicates hallucinations in the reference.}
\label{table:examples_corpora}
\end{table*}

\clearpage

\section{Our Model}
\label{app:model_design}
 \begin{figure}[h]
    \centering
    \includegraphics[width=1.0\linewidth]{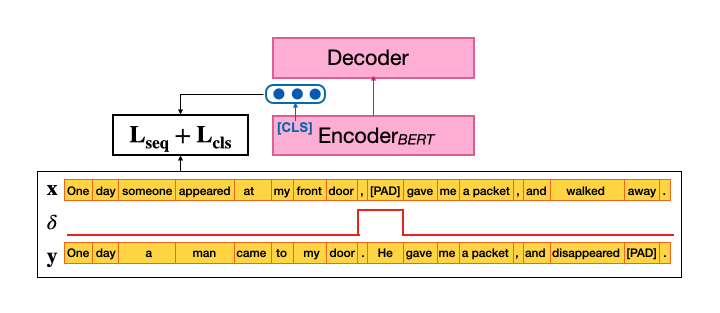}
    \caption{Overview of our proposed approach, where the model is trained on only the split-related edits in $\mathbf{y}$ determined by edit $\mathbf{\delta}$ and sentence category $l$ labels. We also illustrate the padding and the construction of $\mathbf{\delta}$ for a training pair belonging to \textit{Changes Near Split} category.}
    \label{fig:model_architecture}
\end{figure}

\clearpage

\section{Examples of System Outputs}
\label{app:system_outputs}

\begin{table*}[h!]
\setlength{\tabcolsep}{2pt}
\scriptsize
\centering
\begin{tabular}{p{3.4cm}|p{12.6cm}}
\toprule
\textbf{Model w/ Data} & \textbf{System Outputs} \\

\midrule
\multicolumn{2}{c}{\textbf{\textit{Preposition / Subordinate Clause}}} \\
\midrule
Source &  To that end, the annual contribution from each Member State shall be established on the basis of the gross national product \hlblue{(GNP) according} to the scale used for determining the GNP element in own resources for financing the general budget of the European Union. \\
Transformer w/ \textsc{Wiki} &  To that end, the annual contribution from each Member State shall be established on the basis of the gross national product (GNP)\hlred{. According to the scale} used for determining the GNP element in own resources for financing the general budget of the European Union . \\
Transformer w/ \textsc{BiSECT} &  To this end, the annual contribution of each Member State shall be established on the basis of the gross national product (GNP)\hlred{. the scale used to determine} the GNP element in own resources for the financing of the general budget of the European Union . \\
Transformer$_{control}$ w/ \textsc{BiSECT} &  To that end, the annual contribution from each Member State shall be established on the basis of the gross national product\hlgreen{. This depends on the scale} used \hlgreen{to determine} the GNP element in own resources for financing the general budget of the European Union . \\
\midrule
\multicolumn{2}{c}{\textbf{\textit{Gerund}}} \\ 
\midrule
Source & Member States shall take appropriate and practicable measures to reduce the risk of transmission of avian influenza from birds living in the wild to susceptible birds kept in zoos\hlblue{, taking into account} the criteria and risk factors set out in Annex I. \\
Transformer w/ \textsc{Wiki} & Member States shall take appropriate and practicable measures to reduce the risk of transmission of avian influenza from birds living in the wild to susceptible birds kept in zoos\hlred{. Taking into account} the criteria and risk factors set out in Annex I.\\
Transformer w/ \textsc{BiSECT} &Member States shall take appropriate and practical measures to reduce the risk of transmission of avian influenza from birds living in the wild to susceptible birds kept in zoos\hlgreen{. These measures will take into account} the criteria and risk factors set out in Annex I.\\
Transformer$_{control}$ w/ \textsc{BiSECT} & Member States shall take appropriate and practicable measures to reduce the risk of transmission of avian influenza from birds living in the wild to susceptible birds kept in zoos\hlgreen{. They must take into account} the criteria and risk factors set out in Annex I.\\
\midrule
\multicolumn{2}{c}{\textbf{\textit{Concluding Relative Clause}}} \\ 
\midrule
Source & There were two main studies involving a total of 529 patients with SLE\hlblue{, in which} Riquent was compared with a placebo (a dummy treatment).\\
Transformer w/ \textsc{Wiki} & There were two main studies involving a total of 529 patients with SLE\hlred{. In which Riquent was} compared with a placebo (a dummy treatment).\\
Transformer w/ \textsc{BiSECT} & There were two main studies involving a total of 529 patients with SLE\hlgreen{. Riquent was} compared with a placebo (a dummy treatment).\\
Transformer$_{control}$ w/ \textsc{BiSECT} & There were two main studies involving a total of 529 patients with SLE\hlgreen{. Riquent was} compared with a placebo (a dummy treatment).\\
\midrule
\multicolumn{2}{c}{\textbf{\textit{Conjunction Without Subject}}} \\
\midrule
Source &  Undertakings required to maintain separate accounts are undertakings that enjoy special or exclusive rights granted by a Member State as referred to in Article 86 (1) of the Treaty \hlblue{or are} entrusted with the operation of a service of general economic interest as referred to in Article 86 (2) of the Treaty\hlblue{ and receive} State aid in any form whatsoever in relation to such service and that carry on other activities . \\
Transformer w/ \textsc{Wiki} &  Undertakings required to maintain separate accounts are undertakings that enjoy special or exclusive rights granted by a Member State as referred to in Article 86 (1) of the Treaty\hlgreen{. They may} also be entrusted with the operation of a service of general economic interest as referred by the Treaty and receive State aid in any form whatsoever in relation to such service and that carry on other activities . \\
Transformer w/ \textsc{BiSECT} &  Undertakings required to maintain separate accounts are undertakings that enjoy special or exclusive rights granted by a Member State under Article 86 (1) of the Treaty or are entrusted with the operation of a service of general economic interest referred to in Article 86 \hlred{Para . 2) and} receive State aid in any form in relation to this service and carry out other activities . \\
Transformer$_{control}$ w/ \textsc{BiSECT} &  Undertakings required to maintain separate accounts are undertakings that enjoy special or exclusive rights granted by a Member State as referred to in Article 86 (1) of the Treaty or are entrusted with the operation of a service of general economic interest\hlgreen{. They must also} receive state aid in any form whatsoever in relation to such service and that carry on other activities . \\
\midrule
\multicolumn{2}{c}{\textbf{\textit{Preceding Relative Clause}}} \\ 
\midrule
Source & \hlblue{Because 1'-hydroxymidazolam} is an active metabolite\hlblue{, the} sedative effect of midazolam may be increased.\\
Transformer w/ \textsc{Wiki} &\hlred{Because 1'- hydroxymidazolam} is an active metabolite\hlgreen{. The} sedative effect of midazolam may be increased.\\
Transformer w/ \textsc{BiSECT} & \hlgreen{1'-hydroxymidazolam} is an active metabolite\hlgreen{. The} sedative effect of midazolam can be increased.\\
Transformer$_{control}$ w/ \textsc{BiSECT} & \hlgreen{1'-hydroxymidazolam} is an active metabolite\hlgreen{. The} sedative effect of midazolam may therefore be increased.\\
\bottomrule
\end{tabular}
\caption{Examples of system outputs from the \textsc{BiSECT} test set. Here, the source sentence belongs to the category ``\textbf{Changes Near Split}''. \hlblue{Blue} marks the location of the required edits in the source sentence. \hlgreen{Green} indicates good edits and \hlred{red} indicates errors.}
    \label{tab:sample_outputs_category_2}
\end{table*}

\clearpage 

\section{Human Evaluation}
\label{app:human_eval}

\begin{figure*}[h!]
\centering
\includegraphics[width=1.0\linewidth]{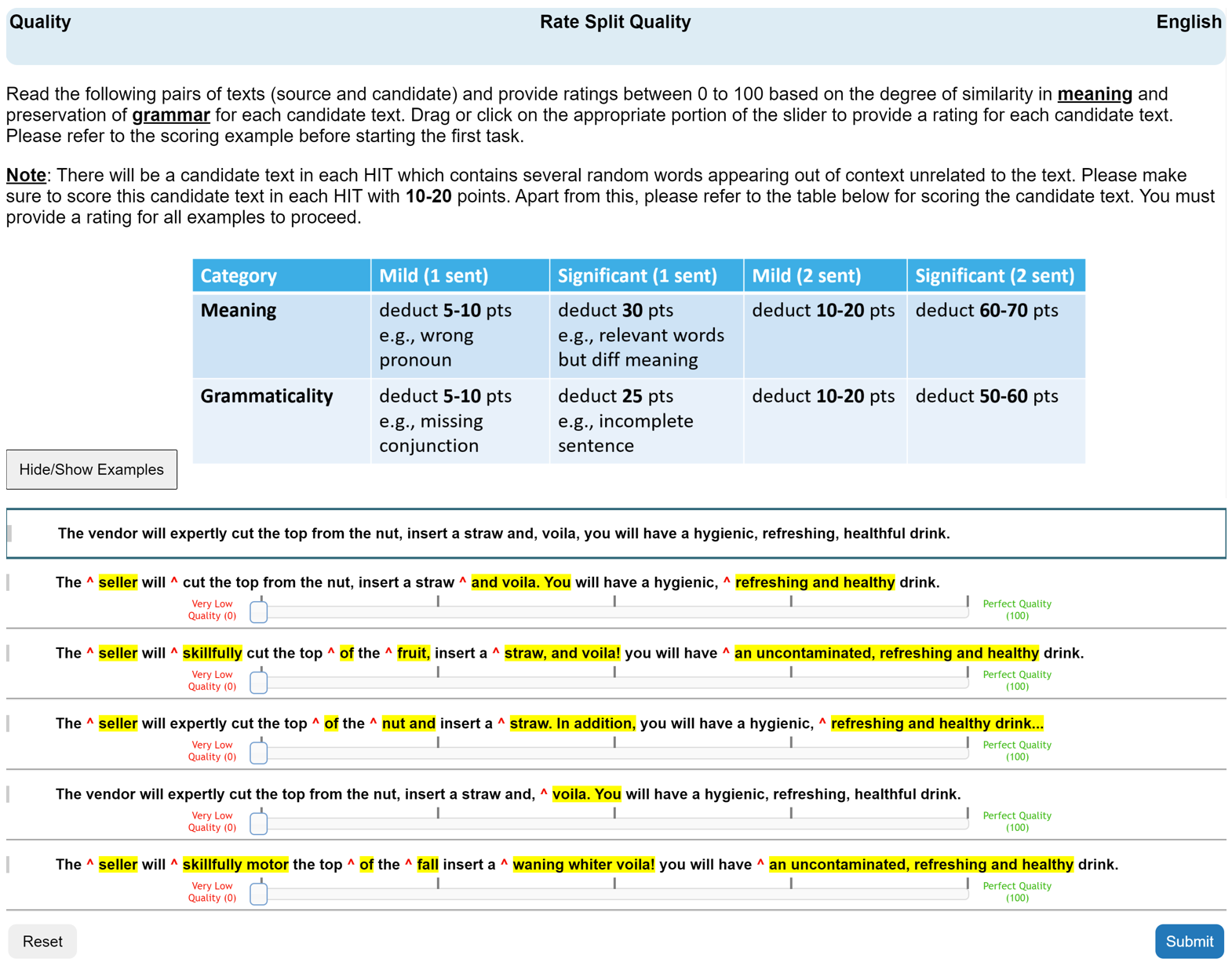}
\caption{Annotation interface and guidelines for human evaluation. 
Each system output is followed by a slider ranging between 0 to 100 with labels ``\textit{Very Low Quality}'' on the left and ``\textit{Perfect Quality}'' on the right. Highlighted words indicate newly added words when compared to the source sentence. Hovering the mouse over the red ticks displays words removed from the source sentence. Every HIT contains a control text, where 4 to 8 words are replaced with random words. Workers are expected to give low scores to the control text. Furthermore, the system outputs are shuffled for every HIT to eliminate position bias.}
\label{fig*:human_eval_design}
\end{figure*}

\clearpage

\section{Multilingual BiSECT}
\label{app:multilingual}

\subsection{French}

\begin{table*}[h!]
\setlength{\tabcolsep}{4pt}
\centering
\small
\begin{tabular}{l|ccrlrcc}
  \toprule
   \multirow{2}{*}{\textbf{Dataset}} & \multirow{2}{*}{\textbf{Pivot Lang.}}  & \multirow{2}{*}{\textbf{Domain}} & \multicolumn{3}{c}{\textbf{1-2 \& 2-1 Alignments} }  & \multicolumn{2}{c}{\textbf{Sent. Length}} \\
   & & & \multicolumn{2}{c}{total (count/\%)} & after filtering  & long & split \\
  \midrule
  \textsc{CCAligned}  & en  & web crawl   & 164,628 & (12.85\%) & 56,799{ }{ }{ }{ }{ } & 37.16 & 41.05\\  
  \textsc{Europarl}  & en &  European Parliament & 153,220 & (11.96\%) & 57,581{ }{ }{ }{ }{ } & 46.30 & 47.74\\
  \textsc{Gigaword} & en & newswire & 624,372 & (48.73\%) & 235,133{ }{ }{ }{ }{ } & 43.73 & 44.58\\
  \textsc{ParaCrawl} & en & web crawl & 308,047 & (24.04\%) & 127,655{ }{ }{ }{ }{ } & 39.07 & 39.03\\
  \textsc{UN} & en & United Nations & 23,706 & (1.85\%) & 13,869{ }{ }{ }{ }{ } & 47.45 & 49.93\\ \cmidrule{1-8}
  \textsc{EMEA}  & en &  European Medicines Agency & 5,719 & (0.45\%) & 2,400{ }{ }{ }{ }{ } & 40.03 & 45.12\\
  \textsc{JRC-Acquis} & en & European Union  & 1,690 & (0.13\%) & 1,036{ }{ }{ }{ }{ } & 49.11 & 52.94\\
  
\bottomrule
\end{tabular}
\caption{Statistics of datasets in the OPUS collection that we used to create French version of the BiSECT corpus.}
\label{tab:BiSECT_French}
\end{table*}

\subsection{Spanish}

\begin{table*}[h!]
\setlength{\tabcolsep}{4pt}
\centering
\small
\begin{tabular}{l|ccrlrcc}
  \toprule
   \multirow{2}{*}{\textbf{Dataset}} & \multirow{2}{*}{\textbf{Pivot Lang.}}  & \multirow{2}{*}{\textbf{Domain}} & \multicolumn{3}{c}{\textbf{1-2 \& 2-1 Alignments} }  & \multicolumn{2}{c}{\textbf{Sent. Length}} \\
   & & & \multicolumn{2}{c}{total (count/\%)} & after filtering  & long & split \\
  \midrule
  \textsc{CCAligned}  & en  & web crawl   & 466,240 & (56.16\%) & 110,958{ }{ }{ }{ }{ } & 40.45 & 46.11\\  
  \textsc{ParaCrawl} & en & web crawl & 297,879 & (35.88\%) & 162,048{ }{ }{ }{ }{ } & 35.36 & 33.75\\
  \textsc{UN} & en & United Nations & 17,948 & (2.16\%) & 9,938{ }{ }{ }{ }{ } & 48.02 & 51.76\\ \cmidrule{1-8}
  \textsc{Europarl}  & en &  European Parliament & 48,165 & (5.80\%) & 6,719{ }{ }{ }{ }{ } & 46.68 & 47.90\\
  
\bottomrule
\end{tabular}
\caption{Statistics of datasets in the OPUS collection that we used to create Spanish version of the BiSECT corpus.}
\label{tab:BiSECT_Spanish}
\end{table*}

\subsection{German}

\begin{table*}[h!]
\setlength{\tabcolsep}{4pt}
\centering
\small
\begin{tabular}{l|ccrlrcc}
  \toprule
   \multirow{2}{*}{\textbf{Dataset}} & \multirow{2}{*}{\textbf{Pivot Lang.}}  & \multirow{2}{*}{\textbf{Domain}} & \multicolumn{3}{c}{\textbf{1-2 \& 2-1 Alignments} }  & \multicolumn{2}{c}{\textbf{Sent. Length}} \\
   & & & \multicolumn{2}{c}{total (count/\%)} & after filtering  & long & split \\
  \midrule
  \textsc{CCAligned}  & en  & web crawl   & 510,817 & (52.57\%) & 52,253{ }{ }{ }{ }{ } & 30.87 & 36.65\\
  \textsc{Europarl}  & en &  European Parliament & 100,784 & (10.37\%) & 16,359{ }{ }{ }{ }{ } & 42.08 & 44.24\\
  \textsc{ParaCrawl} & en & web crawl & 353,136 & (36.34\%) & 116,026{ }{ }{ }{ }{ } & 37.73 & 38.99\\\cmidrule{1-8}
  \textsc{JRC-Acquis} & en & European Union  & 6,950 & (0.72\%) & 1,599{ }{ }{ }{ }{ } & 54.79 & 55.19\\
  
\bottomrule
\end{tabular}
\caption{Statistics of datasets in the OPUS collection that we used to create German version of the BiSECT corpus.}
\label{tab:BiSECT_German}
\end{table*}

\end{document}